\documentclass[10pt,twocolumn,letterpaper]{article}

\usepackage[pagenumbers]{cvpr} % To force page numbers, e.g. for an arXiv version

\usepackage{graphicx}
\usepackage{amsmath}
\usepackage{amssymb}
\usepackage{booktabs}

\usepackage[nolist,nohyperlinks]{acronym}
\usepackage[usenames,dvipsnames]{color}
\usepackage{multirow}
\usepackage[accsupp]{axessibility}

\usepackage[pagebackref,breaklinks,colorlinks]{hyperref}

\usepackage[capitalize]{cleveref}
\crefname{section}{Sec.}{Secs.}
\Crefname{section}{Section}{Sections}
\Crefname{table}{Table}{Tables}
\crefname{table}{Tab.}{Tabs.}

\def\cvprPaperID{23} % *** Enter the CVPR Paper ID here
\def\confName{CVPR}
\def\confYear{2022}

\begin{document}

\begin{acronym}[ICANN]
    \acro{cnn}[CNN]{Convolutional Neural Network}
    \acro{gnn}[GNN]{Graph Neural Network}
    \acro{nn}[NN]{Neural Network}
    \acro{fc}[FC]{fully-connected}
    \acro{lstm}[LSTM]{Long Short-Term Memory}
    \acro{rnn}[RNN]{Recurrent Neural Network}
    \acro{mot}[MOT]{Multi Object Tracking}
    \acro{mota}[MOTA]{Multi Object Tracking Accuracy}
    \acro{motp}[MOTP]{Multi Object Tracking Precision}
    \acro{amota}[AMOTA]{Average Multi Object Tracking Accuracy}
    \acro{amotp}[AMOTP]{Average Multi Object Tracking Precision}
    \acro{hota}[HOTA]{Higher Order Tracking Accuracy}
    \acro{mot}[MOT]{Multi-Object Tracking}
    \acro{iou}[IoU]{Intersection over Union}
    \acro{pd}[PD]{Parallel Domain}
    %% Define a custom plural form of an acronym   
    \acrodefplural{cnn}[CNNs]{Convolutional Neural Networks} % \aclp{cnn}
    \acrodefplural{gnn}[GNNs]{Graph Neural Networks} % \aclp{gnn}
    \acrodefplural{nn}[NNs]{Neural Networks} % \aclp{nn}
\end{acronym}

\title{TripletTrack: 3D Object Tracking using Triplet Embeddings and LSTM}

% \and

\author{%
    \centerline{{Nicola Marinello$^{1}$, Marc Proesmans$^{1,3}$, Luc Van Gool$^{1,2,3}$,}}\\
    \centerline{$^1$ KU Leuven/ESAT-PSI, $^2$ ETH Zurich/CVL, $^3$ TRACE vzw} \\
    \centerline{\small{\texttt{\{nicola.marinello,marc.proesmans,luc.vangool\}@esat.kuleuven.be}}}
}

\maketitle

%%%%%%%%% ABSTRACT
\begin{abstract}
    3D object tracking is a critical task in autonomous driving systems. It plays an essential role for the system's awareness about the surrounding environment. At the same time there is an increasing interest in algorithms for autonomous cars that solely rely on inexpensive sensors, such as cameras. In this paper we investigate the use of triplet embeddings in combination with motion representations for 3D object tracking. We start from an off-the-shelf 3D object detector, and apply a tracking mechanism where objects are matched by an affinity score computed on local object feature embeddings and motion descriptors. The feature embeddings are trained to include information about the visual appearance and monocular 3D object characteristics, while motion descriptors provide a strong representation of object trajectories.  We will show that our approach effectively re-identifies objects, and also behaves reliably and accurately in case of occlusions, missed detections and can detect re-appearance across different field of views. Experimental evaluation shows that our approach outperforms state-of-the-art on nuScenes by a large margin. We also obtain competitive results on KITTI.
\end{abstract}

%-------------------------------------------------------------------------

%%%%%%%%% BODY TEXT
\section{Introduction}
\label{sec:intro}

3D \ac{mot} is a crucial task for autonomous cars since it plays a central role in providing surround situational awareness. 
At the same time there is an increasing interest in exploring perception algorithms that solely rely on cameras since they are relatively inexpensive, compared to other sensors such as LiDARs and radars. Nevertheless camera-only 3D \ac{mot} methods so far have not received as much attention as 2D methods~\cite{motreview}. Recent developments in monocular 3D object detection open new possibilities for 3D object tracking.
The number of 3D object detectors in fact has been increasing recently, demonstrating promising results ~\cite{Simonelli_arXiv_2019,9607436,wang2021probabilistic,NIPS2015_6da37dd3,7780605,8100080,9010867,Shi_2021_ICCV,9607845}. These detectors are capable of estimating the 3D location in camera frame coordinates of an object, along with its size, orientation and category. 3D object detectors allow to work in 3D and thus benefit from the additional information incorporated in the detections. In addition 3D \ac{mot} permits to follow each object in the real world space, a task that is simply not possible with 2D \ac{mot}.

\begin{figure}
    \centering
    \includegraphics[width=\linewidth, trim={0 {5.5cm} 0 {10cm}}, clip]{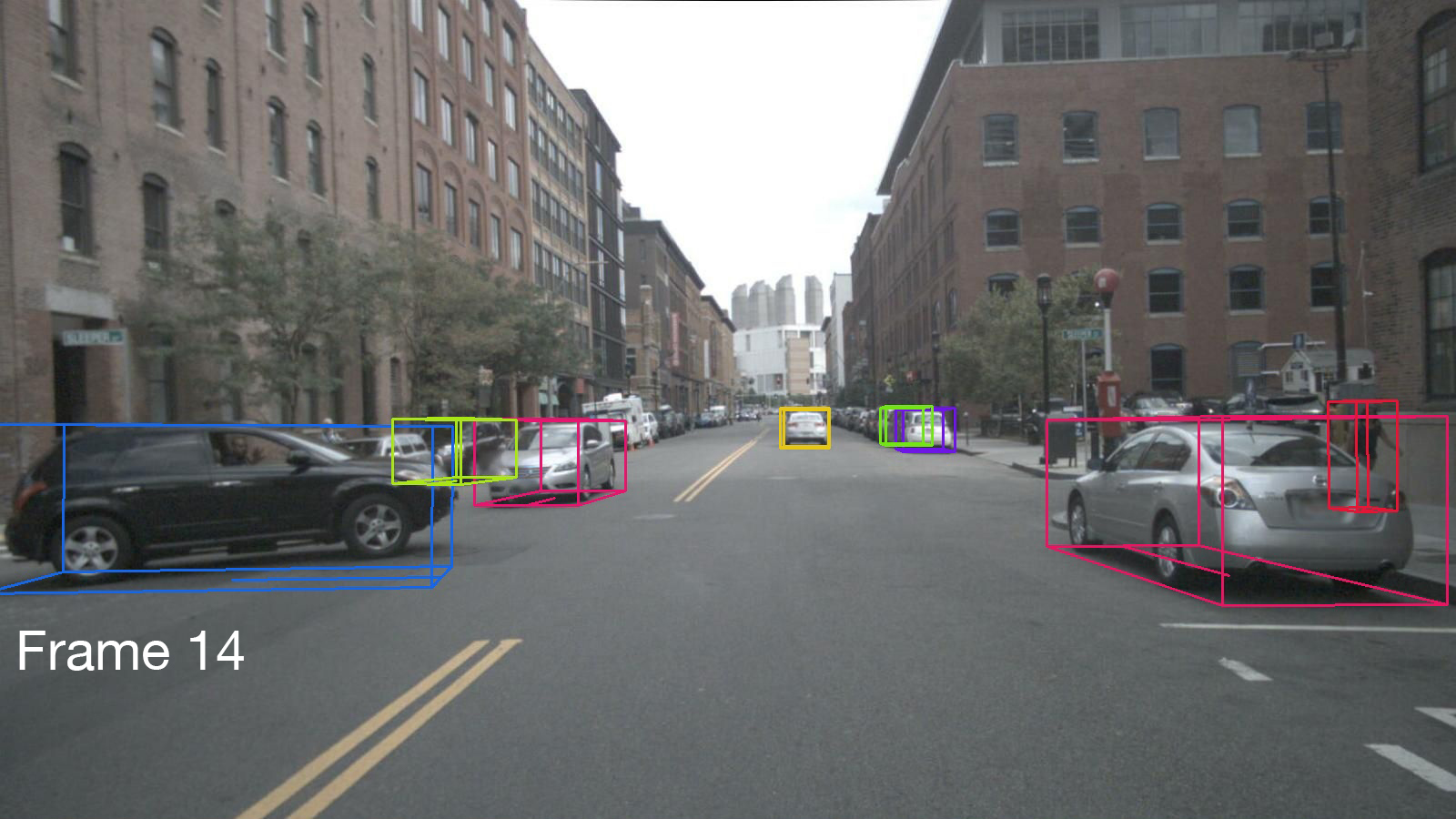}
    \includegraphics[width=\linewidth, trim={0 {5.5cm} 0 {10cm}}, clip]{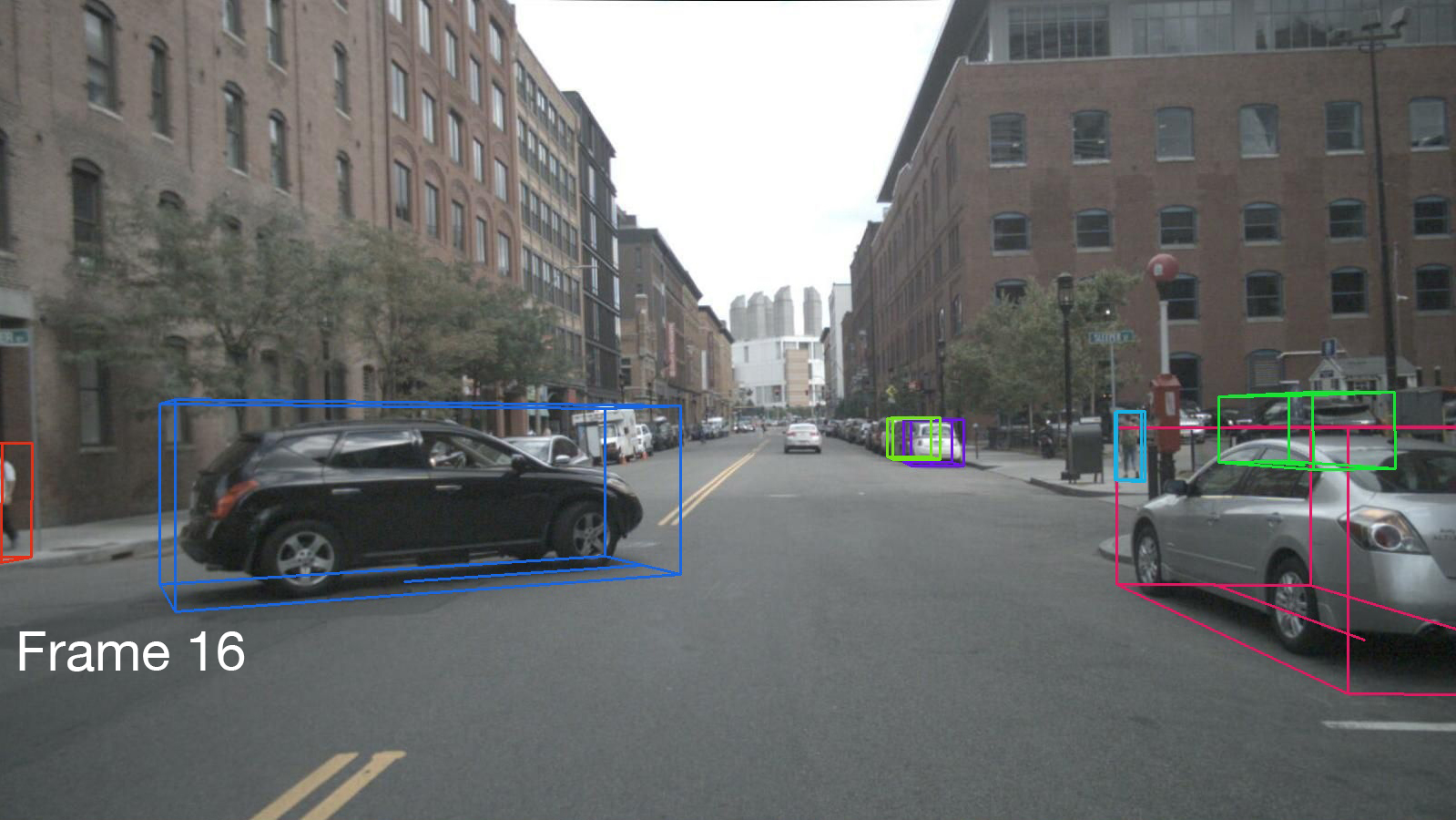}
    \includegraphics[width=\linewidth, trim={0 {5.5cm} 0 {10cm}}, clip]{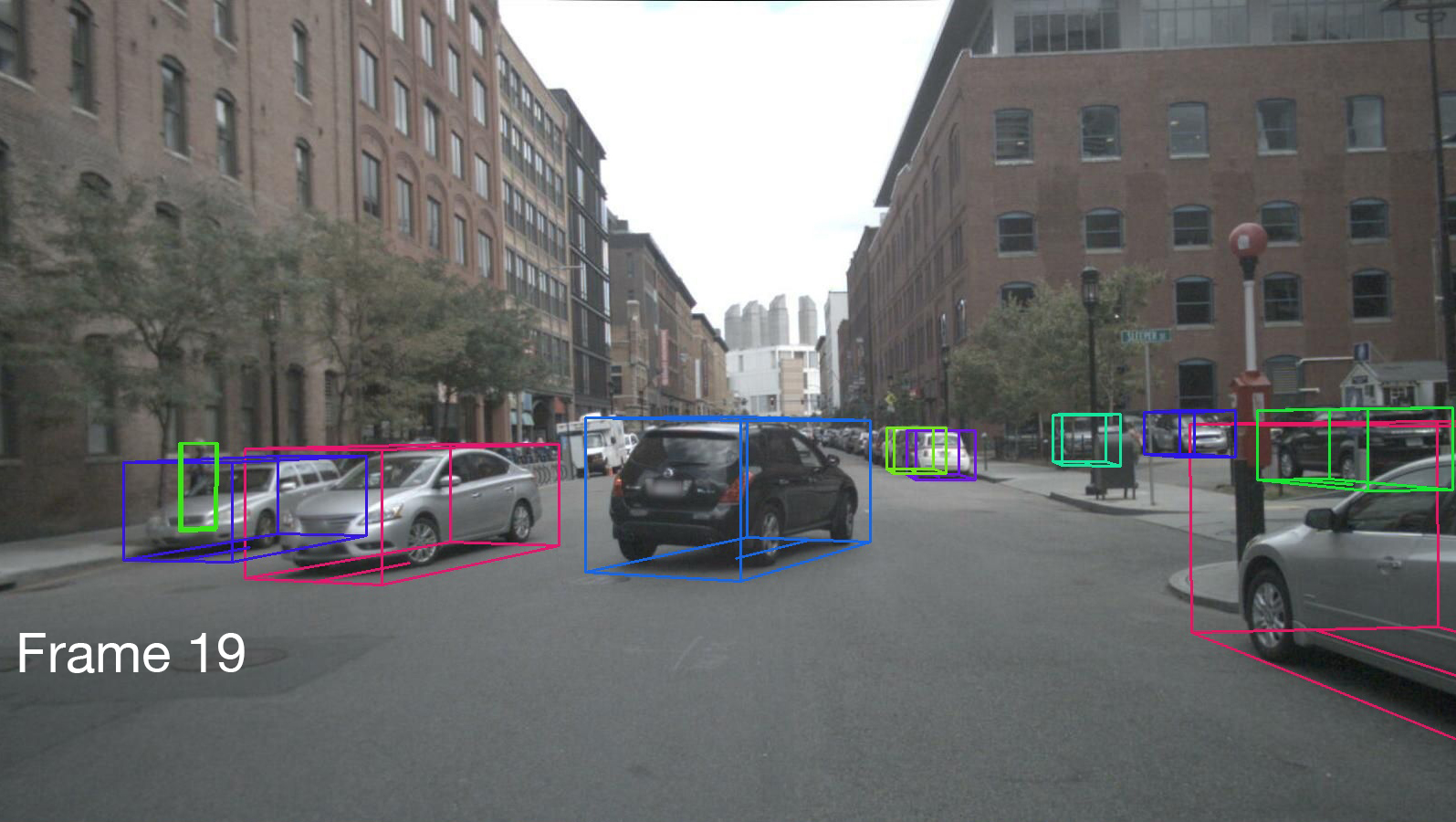}
    \caption{Illustration of re-identification with our tracking pipeline. The fuchsia colored car on the left is being occluded for a few frames by the blue car and the ID is picked up again.
    }
    \label{fig:teaser}
\end{figure}
3D tracking essentially consists of two sub-tasks: 3D detection and re-identification. 3D detection is the operation of recognizing objects in images and estimating their position, dimension and orientation. Re-identification allows to consistently follow each of them and to keep track of their position history. Tracked objects have to be associated with a unique ID and the same ID has to be correctly re-assigned to the same object each time it is detected. Furthermore, the tracking system has to be able to re-assign the same ID to the same object even after a temporary occlusion, a missed detection, or when the object reappears in a different camera.

In this work we introduce an online camera-only tracking mechanism that follows the Tracking-by-Detection paradigm. We operate under the traditional framework that consists of a detection step and a data association step, focusing on the design of the second one. Similarly to previous 2D approaches ~\cite{jiang2019graph,8237303,Mykheievskyi_2020_ACCV} we investigate how to model the motion of the objects in 3D space with a single neural network and how to extract powerful view-independent object visual features~\cite{Mykheievskyi_2020_ACCV}. To extract visual features we borrow techniques from the re-identification ~\cite{hermans2017defense, kumar2019vehicle} and face recognition ~\cite{7298682} literature. Our approach applies an independent detector to all images frames and it tries to link detected objects to trajectories of tracked objects, by exploiting motion and appearance information. Appearance features are condensed in triplet embeddings and the objects motion is modeled through a \ac{lstm}. Identity association is then finally solved with the Hungarian algorithm~\cite{Kuhn55thehungarian} based on an affinity matrix that summarizes affinity scores between previously tracked objects and detected objects. The method can operate with any camera setup and with any off-the-shelf 3D object detector. Through experimental evaluation we show that a traditional approach consisting of a detection step and a data association step can be highly competitive in terms of tracking performance. 

Some state-of-the-art methods perform joint detection and tracking but at the expense of weaker objects representations or lack of synergy between motion and appearance cues. Some also manually design similarity metrics for the data association step, not fully exploiting the power of \aclp{nn} ~\cite{zhou2020tracking,Chaabane2021deft,Hu2021QD3DT,Hu_2019_ICCV}.
We instead advocate for a data-driven method that focuses on strong appearance and motion representations. Exploiting both cues makes it robust in case of temporal occlusions and missed detections, being able to pick up again objects identities. An example is shown in Fig.~\ref{fig:teaser}. Our main contributions can be summarized as follows:

\begin{itemize}
\setlength\itemsep{0em} 
   \item We propose a camera-only tracking method that can be used with any off-the-shelf 3D object detector and any camera setup.
   \item We introduce an explicit embedding model that uses 2D and 3D appearance cues jointly with 3D motion.
   \item We set a new state-of-the-art on the nuScenes camera only tracking benchmark and we obtain competitive results on KITTI.

\end{itemize}

\section{Related Work}
\label{sec:related} 

The 3D tracking task aims at following each detected object and to keep track of their trajectories in the real world space. As for 2D \ac{mot} current research largely focuses on the tracking-by-detection paradigm ~\cite{Weng2020_AB3DMOT_eccvw,pang2021simpletrack,9578166,Zhang_2019_ICCV,Chaabane2021deft}, including state-of-the-art trackers ~\cite{Hu2021QD3DT,Hu_2019_ICCV,9352500,9626850}. ~\cite{Weng2020_AB3DMOT_eccvw} sets a simple baseline using a combination of a 3D Kalman filter and the Hungarian algorithm~\cite{Kuhn55thehungarian}.
QD-3DT ~\cite{Hu2021QD3DT,Hu_2019_ICCV} proposes an unified framework for joint detection, appearance feature extraction and tracking. It combines appearance and predicted motion information to match detections with tracklets. DEFT ~\cite{Chaabane2021deft} also advocates for an unified detection and tracking framework while using a \ac{lstm} that captures motion constraints to filter out physically implausible matches. ~\cite{8813779} integrates a Kalman filter for state estimation and motion prediction along with appearance features to associate detections with tracked objects. ~\cite{9341251} makes use of the triplet loss to learn embeddings that improve data association in a self-supervised manner. On the other side others try to track objects without any appearance information~\cite{Wang_2021_ICCV}. ~\cite{8500454} pairs up an object detector and a recursive filter that predicts and refine object positions and kinematical properties to track objects in 3D space. In \cite{zhou2020tracking} they keep track of objects by consistently following their centers in 2D space, however, they trade the ability to re-identify identities after long occlusions with speed. ~\cite{Tokmakov_2021_ICCV} proposes an approach in 2D that can directly be extended to 3D to hallucinate trajectories of fully occluded objects by extending the CenterTrack model with a recurrent memory module. ~\cite{9362208} performs tracking given point cloud data with and end-to-end network that detects objects and directly assigns tracking IDs. \aclp{gnn} have also been used in the context of 3D \ac{mot} ~\cite{9691806,9157602}. OGR3MOT ~\cite{9691806} models data association, and track management through a graph representation, while GNN3DMOT ~\cite{9157602} extracts appearance and motion features from 2D and 3D spaces and with a \ac{gnn} that models their interaction. ~\cite{MOTBeyondPixels} proposes shape pose and motion costs to improve data association and tracking. ~\cite{luiten2019MOTSFusion} presents a tracking pipeline that exploits 2D and 3D world space motion consistency to improve long-term tracking under occlusions. \cite{benbarka2021score} proposes a confidence-based method for initialization and termination of tracklets. State-of-the-art performance can also be achieved without using the tracking-by-detection paradigm ~\cite{wu2021}. 
Finally, there is an entire line of works that take advantage of multiple sensors by combining their information flows ~\cite{9636498,9562072,9561754,Zhang_2019_ICCV,9636311,9341635}.

\section{Method}
\label{sec:method}

\subsection{TripletTrack}

\begin{figure*}
\centering
\includegraphics[width=1\linewidth]{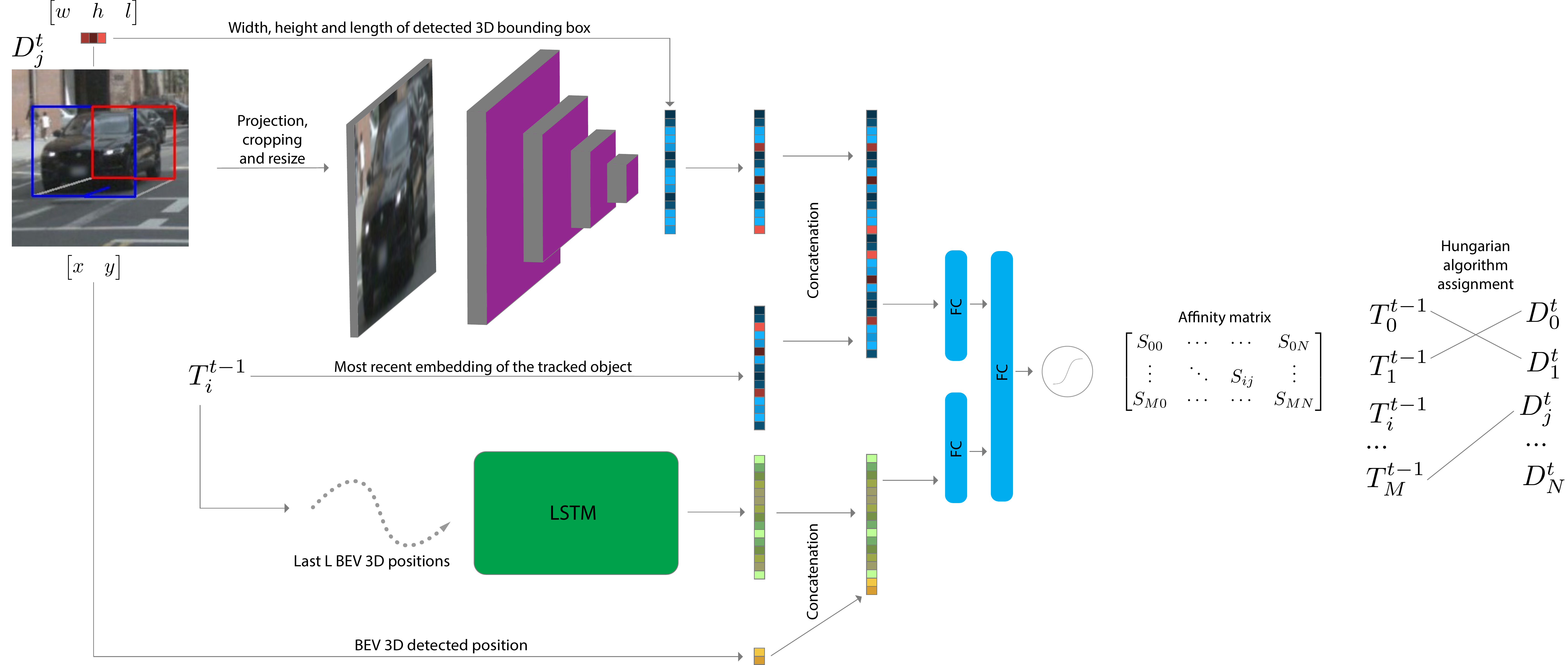}
  \caption{\textbf{Proposed pipeline.} At every time frame $t$ all pairs tracked-detected objects $T_{i}^{t-1}$, $D_{j}^{t}$ are compared to compute an affinity matrix. The computed appearance embedding of a detected object is concatenated with the appearance embedding of a tracked object. The motion representation of a tracked object is concatenated with BEV 3D position of the detected object.
  The appearance and motion components are fed to an affinity network that computes the affinity score for a specific pair of objects.
The affinity matrix is later used by the Hungarian algorithm to associate objects and re-assign the same ID. Lost objects are repeatedly compared with new detections up to a maximum amount of time. If they have not been re-identified after the maximum amount of time they are discarded. Unmatched detections are considered new objects.}
\label{fig:scheme}
\end{figure*}

Let's formalize the re-identification task: we can define the state of a tracked object as $T_{i} = (x, y, z, \theta, w, h, l, \text{ID}, \mathbf{p})$, the state of a detected object $D_{j} = (x, y, z, \theta, w, h, l)$ where $x$, $y$ and $z$ are the center of the 3D bounding box in global coordinates, $\theta$ is the yaw angle, $w$, $h$, and $l$ are the width, height and length respectively, ID is a unique id assigned to a tracked object and $\mathbf{p}$ is the object path history. We can also define the set of tracked objects at time $t-1$ as $T^{t-1}=\{T_{1}^{t-1}, T_{2}^{t-1}, ..., T_{m}^{t-1}\}$ and the set of detected objects $D^{t}=\{D_{1}^{t}, D_{2}^{t}, ..., D_{n}^{t}\}$. The objective of a re-identification algorithm is to take as input the set of previously tracked objects $T^{t-1}$ and the new detections set $D^{t}$ and assign to the detected objects the same ID that they had in the previous frames. If new objects enter into the scene they have to be assigned with a new ID. Human drivers have the innate capacity to perform such a task by implicitly using two cues: \emph{appearance} and \emph{motion}. We therefore take inspiration from this to design an algorithm that exploits these cues.

\textbf{Motion.} Motion can be learned with a Neural Network. The purpose of learning a motion model is to be able to extract a compact representation of an object trajectory in space.
A well suited model for this purpose is the \ac{lstm}. This is a particular type of \ac{rnn} which can process a series of inputs and identify patterns within the input motion sequence. By training the model with objects' trajectories we believe that it can learn how objects typically move in space providing a meaningful representation. Our ablation study in Section \ref{sec:ablation} shows that this can be attained. In this work we aim to train a single \ac{lstm} that can be used to re-identify objects of any class. Since we want to model the objects motion in space, we believe that the most straightforward way is to feed the \ac{lstm} with the 3D position projected on the ground plane, which we will call "BEV 3D position", expressed in global coordinates. This can be achieved by using a 3D object detector and a GPS/IMU that determines the ego pose with respect to a global reference system.

\textbf{2D Appearance.} The other key attribute is the visual appearance of an object, such as an objects shape, or its reflectance properties observed through a camera or set of cameras.
Appearance information is especially desirable in case of long lasting occlusions of dynamic objects. When an object is not in sight for a long time, the motion analysis may struggle to re-identify it. In these cases, appearance information provides a strong clue that allows to correctly assign the ID to the object that reappeared. In order to effectively extract robust object visual features, we employed a dedicated \ac{cnn} that computes triplet embeddings. Ideally, the visual characteristics of an object should be invariant to the view point, scale, illumination, partial occlusions, truncation and background clutter. 

\textbf{3D Appearance.} Each detection out of a 3D object detector comes with the estimated 3D bounding box size. This information can be regarded as an additional appearance feature and exploited to improve the embeddings discriminative ability. We therefore concatenate the estimated 3D bounding box size of the corresponding object at the last layer of the \ac{cnn}.

\textbf{Affinity network.}
Our re-identification method extracts appearance features with the \ac{cnn}, and motion features with the \ac{lstm} from tracked and detected objects. A tiny affinity network is then responsible to combine them and to determine their similarity. The re-identification pipeline is shown in Fig.~\ref{fig:scheme}.

\subsection{Training}

The proposed pipeline is trained in two stages, and during the entire training process we make use of the annotations provided with the dataset. This makes the training process completely independent from the object detector that will be used during inference time.

In the first stage we train the \ac{cnn} to learn objects embeddings. We use the \emph{triplet loss}, first proposed in~\cite{7298682}. The triplet loss has been extensively used in the past, especially in the context of re-identification for surveillance purposes ~\cite{hermans2017defense,8578937,8578227,kumar2019vehicle}. It is computed over a triplet of images: an \emph{anchor}, a \emph{positive} and a \emph{negative}; the anchor and the positive images show the same object, while the negative shows an object with a different identity. It learns an embedding space by forcing embeddings of the same identity to be closer to each other than an embedding from any other identity by at least a margin $m$. Since the aforementioned datasets provide unique identity annotations for each object, they can be used as ground truth labels to train an embedding model with the triplet loss. The triplet loss is expressed as follows:

\begin{equation}
\mathcal{L}_{\text {tri }}=\sum_{a, p, n}\left[m+D_{a, p}-D_{a, n}\right]_{+}
\label{eq:triplet}
\end{equation}
where $D$ is the embedding distance. 
As input to the network we provide crops of projected ground truth 3D bounding boxes and the corresponding 3D bounding box sizes. A triplet of crops can contain objects from any scene at any frame in the dataset. We supervise the network with the triplet loss and the ground truth objects identity labels from the dataset. However, particularly in automotive datasets that provide continuous camera recordings, the similarity between the same object in succeeding frames is very high, hence it is crucial to train the \ac{cnn} on the most informative triplets. Therefore we followed the online mining strategy proposed by~\cite{hermans2017defense}. The training batches are built in the following way: $P$ identities are randomly sampled from the dataset and then $K$ images of each identity to build batches of $PK$ images. Over such a batch, a maximum of $PK(PK-K)(K-1)$ triplets combinations can potentially be sampled. In order to sample the most informative triplets we use the \emph{batch hard} online mining~\cite{hermans2017defense} to compute the triplet loss over the hardest triplets, resulting in $PK$ triplet contributions in the loss computation.

In the second stage we jointly train the entire pipeline that includes the \ac{lstm}, the affinity network and the \ac{cnn}. However the \ac{cnn} weights are not updated in this stage.  The pipeline is trained in batches and each batch element corresponds to a specific frame of a specific scene in the dataset. A single batch contains frames from any scene and in random order. For each frame a maximum amount of tracked objects and detected objects is randomly sampled. With a tracked object we refer to any object whose most recent annotation in the scene was there at most $B$ frames in the past. With a detected object we refer to any annotated object in the current frame. We supervise the pipeline using the \emph{binary cross entropy loss} and a ground truth affinity matrix. Let $M$ denote the number of sampled tracked objects and $N$ the number of sampled detected objects in a specific frame. We define a ground truth affinity matrix $Y\in\{0,1\}^{M \times N}$ using their identity labels. For every pair $(T_{i}^{t-1}, D_{j}^{t})$ the pipeline determines if they correspond to the same identity by computing a similarity score. The affinity network receives as input the appearance embedding\footnote{The appearance embedding of a tracked object is computed over its last visible annotation.} and the position history representation of $T_{i}^{t-1}$, along with the appearance embedding and the BEV 3D position of $D_{j}^{t}$.  These cues are combined and used to compute the estimated affinity $\hat{y_{i j}}\in[0,1]$. By repeating this process for every pair $(T_{i}^{t-1}, D_{j}^{t})$ we obtain an affinity matrix $\hat{Y}\in[0,1]^{M \times N}$. To produce the feature embeddings the CNN is fed with the crop of the projected 3D bounding box and the 3D bounding box size. While the \ac{lstm} processes the BEV 3D position history of the tracked object. The binary cross entropy loss is then computed over corresponding elements of the two matrices as follows:

\begin{equation}
\mathcal{L}_{e}=\sum_{i}^{M} \sum_{j}^{N} (-p \ y_{ij} \log(\hat{y_{i j}}) + (1-y_{i j})\log(1-\hat{y_{i j}}))
\label{eq:bce}
\end{equation}
where $y_{ij}\in Y$, $\hat{y_{i j}} \in \hat{Y}$ and p is the weight for the positive pairs. The ground truth $y_{ij}$ corresponds to $1$ when the two identity labels are the same, $0$ otherwise. Moreover, at each frame, the number of positive pairs is much lower than the number of negative pairs, hence, when computing the binary cross entropy loss, we balance the positive weight $p$ in the loss accordingly. 

However BEV 3D coordinates cannot be directly fed to the \ac{lstm} as they are. They can in fact become arbitrarily large, depending on the distance between the object and the origin of the reference system. The consequence is that the training process is numerically unstable. To overcome this problem we translate each motion sequence point w.r.t the first point of the input sequence. We also limit the number of last positions processed by the \ac{lstm} to a maximum of $L$. With these tricks the input values always span approximately the same range and let the training kick off. The BEV 3D position of $D_{j}^{t}$ is also given w.r.t. the first $T_{i}^{t-1}$ sequence point.

\subsection{Inference}

During inference time, the pipeline is run over sequences and it takes as input detections from the object detector. At each frame it computes the similarity score between previously tracked objects and detected objects in the current frame. The estimated affinity matrix is then solved by the Hungarian algorithm~\cite{Kuhn55thehungarian} to complete the assignment task. The lowest tolerated score for a match is set to $0.5$. Solving the affinity matrix with the Hungarian algorithm implies to find the pairs that are likely to be the same object, based on their motion and their aspect. The process is repeated frame-by-frame to re-identify objects over time.

\textbf{Tracking score}. A tracking score is given to each detected object after the pipeline processing. To compute a value that takes into account detection and re-identification, we compute the tracking score as the product between the detection score and the affinity score. When a detected object is considered to be new we set the tracking score equal to the detection score.

\section{Implementation and evaluation}
\label{sec:impl-eval}

\subsection{Datasets and metrics}

\textbf{nuScenes.} nuScenes~\cite{nuscenes2019} is a multimodal autonomous driving dataset released in 2019. It contains annotated sequences of approximately 20 seconds long captured from a vehicle driving in urban environments. The dataset provides data acquired from different sensors namely cameras, LiDAR and radars. The number of cameras is 6 and they capture the full 360 degrees view around the car. Different types of objects are annotated with 3D bounding boxes in global coordinates at a frame rate of 2 FPS. These frames are named \emph{keyframes}. Each annotated object is provided with a unique ID across the entire dataset which in our method serves as ground truth supervision. nuScenes also provides frames in between annotated frames, making available a stream of 12 FPS. However, since our method relies on annotations during the training processes, we only used keyframes during the evaluation. nuScenes hosts a 3D tracking benchmark, making it suitable to evaluate our method. 

\textbf{KITTI.} KITTI \ac{mot} ~\cite{Geiger2012CVPR} is a single-camera \ac{mot} benchmark. It includes 21 training sequences and 29 test sequences of footage recorded by a single camera, mounted on top of a car. Several object categories are annotated with 3D bounding boxes in camera coordinates and unique IDs, similarly to nuScenes. GPS/IMU data is also available, which allows objects to be expressed w.r.t. a global coordinates system.

\textbf{Metrics.} To evaluate the proposed tracking method we rely on the evaluation tools provided by nuScenes and KITTI. 

The nuScenes 3D tracking benchmark uses evaluation metrics introduced in~\cite{Weng2020_AB3DMOT_eccvw}, named \ac{amota} and \ac{amotp}. These modified metrics have been introduced to address the main limitation of the conventional CLEAR metrics \ac{mota} and \ac{motp}~\cite{Bernardin2008EvaluatingMO}. \ac{mota} accounts for all object configuration errors made by the tracker: false positives, misses, mismatches, over all frames. \ac{motp} shows the ability of the tracker to estimate precise object positions, independent of its skill at recognizing object configurations and keeping consistent trajectories. Conventional metrics are computed at the best recall value that yields the maximum value of \ac{mota}. However in this way the metrics do not provide an exhaustive overview of the tracker performance at different recall values. To overcome this issue the \ac{amota} and \ac{amotp} are computed as the integral of \ac{mota} and \ac{motp} across the spectrum of recall values, better summarizing the tracker performance. 

KITTI \ac{mot} benchmark assesses the performance of a tracker using the recently proposed \ac{hota} ~\cite{Luiten2020IJCV}. This metric is computed as the combination of three \ac{iou} scores that account for localization, detection and association. For this reason a single number can be used to rank tracker performance. The \ac{hota} metric allows to focus on single abilities of a tracker and to compare them in a more fine-grained manner e.g. it extends the concept of precision and recall to measure association performance.

\subsection{Implementation details}

To train the \ac{cnn} we use a batch size of $128$ with $K=4$ and $P=32$. The margin value $m$ in the triplet loss is set to 1. Using a hard mining strategy (in particular selecting the hardest negatives) to train the \ac{cnn} could lead to a collapsed model i.e. the network predicts every embedding close to $0$ ~\cite{7298682}. This problem can be solved by first warming up the network by sampling \emph{semi-hard} triplets as proposed in ~\cite{7298682} or  with the \emph{batch all} online mining strategy~\cite{hermans2017defense} and then continue the training with the batch hard mining. Note also that constructing batches with two or more identical vehicles with different label identities, might potentially lead to distortions in the embedding space. In fact they would be pushed far apart by the triplet loss since they would be treated as negative examples. Nevertheless, we assume the probability of two or more identical vehicles, ending up in the same batch sufficiently low.
For our experiments we use a ResNet-18~\cite{7780459} with pre-trained weights on ImageNet~\cite{conf/cvpr/DengDSLL009} where we replaced the last layer with a \ac{fc} one of size $128$. As optimizer we use Adam~\cite{Adam} with learning rate $10^{-5}$ on nuScenes and $10^{-6}$ when fine-tuning on KITTI.  

\textbf{Crops preparation}. The quality of crops taken out from 3D bounding box projections can vary greatly. In fact, projected boxes would hardly include the target, in case of large occlusions, heavy truncation and projections that lie on the edges of the image. Low quality crops can potentially reduce the quality of the embeddings and introduce distortions in the embedding space. To mitigate this issue we apply filters on the selection of the projected bounding boxes, such that the \ac{cnn} is trained on cleaner crops.
Filtering is carried out based on bounding box resolution, their aspect ratio and visibility score. Crops are resized to a fixed size of $224\times224$.

\textbf{Crops and bounding box sizes augmentation}. To make the network more robust under a variety of conditions, we applied different data augmentation techniques on the input crops. Color jittering, random horizontal flipping, random rotation and cutout~\cite{zhong2017random}. To also make it more robust against fluctuations of estimated 3D bounding box sizes, typical of object detectors, we add noise on width, height and length. We apply a percentage of noise to each dimension of the bounding box drawn from a uniform distribution $X \sim \mathcal{U}(-a,a)$ with $a=0.2$. Adding noise to the 3D bounding box size is necessary. This is due to the fact that the annotated size of an object is consistent over different annotations. In fact, without noise, the size would act as fingerprint that artificially differentiate objects, leading to a network that only inspects the bounding box size to distinguish identities.

To train the second stage we use a batch size of $32$ on nuScenes and $16$ on KITTI. The number of sampled tracked objects $M$ and sampled detected objects $N$ is set to $16$. We set $B$ to a value that approximates $5$ seconds and $L=40$. For our experiments we use a one-layer \ac{lstm} with hidden state size $128$. The affinity network consists of 2 \ac{fc} layers that concatenates the appearance and motion embeddings before passing them to the subsequent layers. In the last layer a sigmoid activation function computes the affinity score. We optimized the weights using Adam~\cite{Adam} with learning rate $5\times10^{-4}$ on nuScenes and $10^{-5}$ when fine-tuning on KITTI.

\textbf{Trajectories data augmentation}. 3D object detectors reliably estimates the boundaries of detected objects. Nevertheless, in particular monocular object detectors, struggle to precisely estimate the object position, introducing more noise in the direction of the camera. For this reason we introduce noise on the position history of tracked objects during training. Even though more noise should be added in the direction of the camera, we add position noise uniformly in every direction for simplicity reasons. The noise added on each position point is drawn from $X \sim \mathcal{N}(\mu,\sigma^{2})$ with $\mu=0$ and $\sigma^{2}=1$. Note that for the aforementioned reason the noise solely affects the 3D objects position and it is not propagated in the projected 2D crop location.

\textbf{Object detector}. We use an existing 3D object detector from the current leading camera-only 3D tracking method on nuScenes to fairly compare with them. We apply our tracking pipeline on raw detections from the QD-3DT~\cite{Hu2021QD3DT} object detector which is based on Faster RCNN~\cite{NIPS2015_14bfa6bb}. The object detector is trained on our target datasets and it performs predictions on monocular images. To facilitate future comparisons and to demonstrate that our method can work with any object detector we also evaluate our pipeline on the validation set of nuScenes using the monocular Mapillary 3D object detector~\cite{Simonelli_arXiv_2019}, which is provided as baseline for the 3D tracking task.

Detections provided by the 3D object detector are filtered by a score threshold. The threshold is set to $0.8$ for every object, except for the pedestrian category on KITTI set to $0.85$. Moreover, since a single object might be visible from multiple cameras, at each time frame, multiple detections for that object might be available. We then apply a non-maxima suppression on duplicate detections based on their 2D \ac{iou}. When an object is visible in multiple cameras, a feature embedding is extracted from all the views. The final object embedding is then computed as the weighted average of the views embeddings by their resolution.

\begin{table}[]
\centering
\resizebox{0.7\columnwidth}{!}{\begin{tabular}{lcc}
\toprule
Method                                       & AMOTA $\uparrow$                     & AMOTP $\downarrow$   \\
\bottomrule
3D center dist                       & $0.208$                        & $1.501$              \\
Our motion model                      & $\mathbf{0.251}$               & $\mathbf{1.489}$              \\
\end{tabular}}
\caption{\textbf{Motion model efficacy.} Experiment on nuScenes validation set to demonstrate \ac{lstm} + affinity network effectiveness. In both cases we use the same set of detections from the Mapillary~\cite{Simonelli_arXiv_2019} 3D object detector.}
\label{tab:nusc-val-motion-abl}
\end{table}

\begin{table}[]
\centering
\resizebox{\columnwidth}{!}{\begin{tabular}{lcc}
\toprule
Method                                                         & AMOTA $\uparrow$                        & AMOTP $\downarrow$   \\
\bottomrule
DEFT ~\cite{Chaabane2021deft}                                                          & $0.209$                                      & $-$              \\
QD-3DT~\cite{Hu2021QD3DT}                                                          & $0.242$                                      & $1.518$              \\
Our appearance model w/ QD-3DT det                      & $0.235$                             & $1.512$              \\
Our motion model w/ QD-3DT det                      & $\mathbf{0.285}$                             & $\mathbf{1.485}$              \\
\end{tabular}}
\caption{\textbf{Our motion model and appearance model vs SOTA.} Experiment on nuScenes validation set. We use QD-3DT~\cite{Hu2021QD3DT} 3D object detector.}
\label{tab:nusc-val-motion-abl-sota}
\end{table}

% % big version
\begin{table*}[ht!]
\centering 
\setlength{\tabcolsep}{1pt}
\resizebox{\textwidth}{!}{\begin{tabular}{lccccccccccccccccc} 
\toprule
Method                                              & AMOTA $\uparrow$          & AMOTP $\downarrow$ & MOTAR & MOTA & MOTP & RECALL & MT & ML & FAF & TP & FP & FN & IDS & FRAG & TID & LGD\\
\bottomrule
CenterTrack-Vision ~\cite{zhou2020tracking}         & $0.046$                   & $1.543$           & $0.046$ & $0.043$ & $0.753$ & $0.233$ & $573$ & $5235$ & $75.945$ & $26544$ & $17574$ & $89214$ & $3807$ & $\mathbf{2645}$ & $2.057$ & $3.819$\\
PermaTrack ~\cite{Tokmakov_2021_ICCV}               & $0.066$                   & $\mathbf{1.491}$           & $0.321$ & $0.060$ & $\mathbf{0.724}$ & $0.189$ & $652$ & $5065$ & $\mathbf{51.342}$ & $29662$ & $\mathbf{16318}$ & $86305$ & $3598$ & $2656$ & $2.163$ & $4.248$\\
DEFT ~\cite{Chaabane2021deft}                       & $0.177$                   & $1.564$           & $0.484$ & $0.156$ & $0.770$ & $0.338$ & $1951$ & $3232$ & $67.741$ & $52099$ & $22163$ & $60565$ & $6901$ & $3420$ & $1.600$ & $3.080$\\
QD-3DT ~\cite{Hu2021QD3DT}                           & $0.217$                   & $1.550$           & $0.563$ & $0.198$ & $0.773$ & $0.375$ & $1893$ & $2970$ & $53.795$ & $52553$ & $16495$ & $60156$ & $6856$ & $3001$ & $1.620$ & $2.961$\\
\bottomrule
\textbf{TripletTrack (Ours)}                        & $\mathbf{0.268}$          & $1.504$  & $\mathbf{0.605}$ & $\mathbf{0.245}$ & $0.800$ & $\mathbf{0.400}$ & $\mathbf{2085}$ & $\mathbf{2922}$ & $60.0$ & $\mathbf{62355}$ & $18517$ & $\mathbf{56166}$ & $\mathbf{1044}$ & $3978$ & $\mathbf{1.33}$ & $\mathbf{2.50}$\\
\end{tabular}}
\caption{\textbf{nuScenes tracking leaderboard}. Tracking leaderboard of nuScenes tracking benchmark for camera-only methods.}
\label{tab:nusc-test}
\end{table*}

%large version -- CAR
\begin{table*}[h!]
\centering
\resizebox{0.6\textwidth}{!}{\begin{tabular}{lcccccccc}
\toprule
Method & HOTA $\uparrow$  & DetA & AssA & DetRe & DetPr & AssRe & AssPr & LocA \\%& MOTA \\
\bottomrule
MOTBeyondPixels ~\cite{MOTBeyondPixels}             & 63.75 \% & 72.87 \% & 56.40 \% & 76.58 \% & 85.38 \% & 59.05 \% & 86.70 \% & 86.90 \% \\
JCSTD ~\cite{8621602}                               & 65.94 \% & 65.37 \% & 67.03 \% & 68.49 \% & 82.42 \% & 71.02 \% & 82.25 \% & 84.03 \% \\
MASS ~\cite{8782450}                                & 68.25 \% & 72.92 \% & 64.46 \% & 76.83 \% & 85.14 \% & 72.12 \% & 81.46 \% & 86.80 \% \\
Quasi-Dense ~\cite{Pang_2021_CVPR}                  & 68.45 \% & 72.44 \% & 65.49 \% & 76.01 \% & 85.37 \% & 68.28 \% & 88.53 \% & 86.50 \% \\
SRK\_ODESA ~\cite{Mykheievskyi_2020_ACCV}           & 68.51 \% & 75.40 \% & 63.08 \% & 78.89 \% & 86.00 \% & 65.89 \% & 87.47 \% & 86.88 \% \\
IMMDP ~\cite{Xiang_2015_ICCV}                       & 68.66 \% & 68.02 \% & 69.76 \% & 71.47 \% & 83.28 \% & 74.50 \% & 82.02 \% & 84.80 \% \\
MOTSFusion ~\cite{luiten2019MOTSFusion}             & 68.74 \% & 72.19 \% & 66.16 \% & 76.05 \% & 84.88 \% & 69.57 \% & 85.49 \% & 86.56 \% \\
TuSimple ~\cite{Choi_2015_ICCV}                     & 71.55 \% & 72.62 \% & 71.11 \% & 76.78 \% & 83.84 \% & 74.51 \% & 86.26 \% & 85.72 \% \\
SMAT ~\cite{10.1007/978-3-030-50516-5_5}            & 71.88 \% & 72.13 \% & 72.13 \% & 74.43 \% & \textbf{87.33 \%} & 74.77 \% & 88.30 \% & 87.19 \% \\
TrackMPNN ~\cite{rangesh2101trackmpnn}              & 72.30 \% & 74.69 \% & 70.63 \% & 80.02 \% & 83.11 \% & 73.58 \% & 87.14 \% & 86.14 \% \\
QD-3DT ~\cite{Hu2021QD3DT}                           & 72.77 \% & 74.09 \% & 72.19 \% & 78.13 \% & 85.48 \% & 74.87 \% & 89.21 \% & 87.16 \% \\
CenterTrack-Vision ~\cite{zhou2020tracking}         & 73.02 \% & 75.62 \% & 71.20 \% & 80.10 \% & 84.56 \% & 73.84 \% & 89.00 \% & 86.52 \% \\
LGM ~\cite{Wang_2021_ICCV}                          & 73.14 \% & 74.61 \% & 72.31 \% & 80.53 \% & 82.16 \% & 76.38 \% & 84.74 \% & 85.85 \% \\
mono3DT ~\cite{Hu_2019_ICCV}                        & 73.16 \% & 72.73 \% & 74.18 \% & 76.51 \% & 85.28 \% & 77.18 \% & 87.77 \% & 86.88 \% \\
DEFT ~\cite{Chaabane2021deft}                       & 74.23 \% & 75.33 \% & 73.79 \% & 79.96 \% & 83.97 \% & 78.30 \% & 85.19 \% & 86.14 \% \\
Mono 3D KF ~\cite{9626850}                          & 75.47 \% & 74.10 \% & 77.63 \% & 78.86 \% & 82.98 \% & 80.23 \% & 88.88 \% & 85.48 \% \\
PermaTrack ~\cite{Tokmakov_2021_ICCV}               & \textbf{78.03 \%} & \textbf{78.29 \%} & \textbf{78.41 \%} & \textbf{81.71 \%} & 86.54 \% & \textbf{81.14 \%} & 89.49 \% & 87.10 \% \\
\bottomrule
\textbf{TripletTrack (Ours)}           & 73.58 \% & 73.18 \% & 74.66 \% & 76.18 \% & 86.81 \% & 77.31 \% & \textbf{89.55 \%} & \textbf{87.37 \%} \\
\end{tabular}}
\caption{\textbf{KITTI \ac{mot} (Car)}. Tracking leaderboard of KITTI \ac{mot} (car) tracking benchmark for camera-only methods.}
\label{tab:kitti-test-car}
\end{table*}

% large version -- PEDESTRIAN
\begin{table*}[h!]
\centering
\resizebox{0.6\textwidth}{!}{\begin{tabular}{lcccccccc}
\toprule
Method & HOTA $\uparrow$  & DetA & AssA & DetRe & DetPr & AssRe & AssPr & LocA \\%& MOTA \\
\bottomrule
TrackMPNN ~\cite{rangesh2101trackmpnn}              & 39.40 \% & 44.24 \% & 35.45 \% & 50.78 \% & 64.58 \% & 38.98 \% & 69.80 \% & 77.56 \% \\
JCSTD ~\cite{8621602}                               & 39.44 \% & 34.20 \% & 45.79 \% & 36.15 \% & 69.39 \% & 49.38 \% & 69.00 \% & 76.23 \% \\
CenterTrack-Vision ~\cite{zhou2020tracking}         & 40.35 \% & 44.48 \% & 36.93 \% & 49.91 \% & 66.83 \% & 41.05 \% & 70.19 \% & 77.81 \% \\
QD-3DT ~\cite{Hu2021QD3DT}                           & 41.08 \% & 44.01 \% & 38.82 \% & 48.96 \% & 67.19 \% & 42.09 \% & 72.44 \% & 77.38 \% \\
Quasi-Dense ~\cite{Pang_2021_CVPR}                  & 41.12 \% & 44.81 \% & 38.10 \% & 48.55 \% & 70.39 \% & 41.02 \% & 72.47 \% & 77.87 \% \\
Mono 3D KF ~\cite{9626850}                          & 42.87 \% & 40.13 \% & 46.31 \% & 46.02 \% & 59.91 \% & \textbf{52.86 \%} & 63.50 \% & 74.03 \% \\
SRK\_ODESA ~\cite{Mykheievskyi_2020_ACCV}           & 43.73 \% & \textbf{53.73 \%} & 36.05 \% & \textbf{58.01 \%} & \textbf{73.19 \%} & 40.05 \% & 69.44 \% & \textbf{78.91 \%} \\
TuSimple ~\cite{Choi_2015_ICCV}                     & 45.88 \% & 44.66 \% & \textbf{47.62 \%} & 47.92 \% & 69.51 \% & 52.04 \% & 69.88 \% & 76.43 \% \\
PermaTrack ~\cite{Tokmakov_2021_ICCV}               & \textbf{48.63 \%} & 52.28 \% & 45.61 \% & 57.40 \% & 71.03 \% & 49.63 \% & \textbf{73.28 \%} & 78.57 \% \\
\bottomrule
\textbf{TripletTrack (Ours)}           & 42.77 \% & 39.54 \% & 46.54 \% & 41.97 \% & 71.91 \% & 50.86 \% & 71.26 \% & 77.93 \% \\
\end{tabular}}
\caption{\textbf{KITTI \ac{mot} (Pedestrian)}. Tracking leaderboard of KITTI \ac{mot} (pedestrian) tracking benchmark for camera-only methods.}
\label{tab:kitti-test-ped}
\end{table*}

\section{Experiments}
\label{sec:experiments}

In this section we will provide ablation experiments and the results that we obtained on the public benchmarks nuScenes and KITTI.

\subsection{Ablation study}
\label{sec:ablation}

\textbf{\ac{lstm} motion model}. We conducted an experiment to verify that the \ac{lstm} has actually learned how objects typically move in space and that it can produce a meaningful representation of object trajectories. We establish as baseline a simple algorithm that re-identifies objects solely based on their 3D bounding box center distance, matching objects by minimizing the sum of the distances between the set of detected objects and the set of tracked objects, with the Hungarian algorithm. We set the maximum tolerated distance for a match to $10$ meters\footnote{$10$ meters is the value that maximizes the metrics performance.}. We compare in Table ~\ref{tab:nusc-val-motion-abl} this baseline with our method trained to exclusively make use of the \ac{lstm} motion model. In this case the affinity network has to be slightly modified since there are no appearance embeddings as input. As can be seen from the table, the \ac{lstm} jointly trained with the affinity network offers superior performance, demonstrating that it is producing a meaningful representation of the objects trajectories instead of merely providing an indication on the last trajectory position.
We also compare in Table~\ref{tab:nusc-val-motion-abl-sota} our motion model with state-of-the-art methods on nuScenes. Surprisingly, our motion model is sufficient to outperform existing methods, providing an additional evidence of the internal \ac{lstm} representation effectiveness.

\textbf{\ac{cnn} appearance model}. Similar to the motion cue we compare our appearance model against state-of-the-art methods on nuScenes. Also in this case a slightly modified affinity network has been trained jointly with the \ac{cnn}. We show in Table~\ref{tab:nusc-val-motion-abl-sota} that our appearance model can compete with or even outperform recent methods.  

To verify that our pipeline benefits from %the usage of 
using appearance features and that the bounding box size increases the quality of the triplet embeddings, we summarize the effect of each cue in Table~\ref{tab:nusc-val-abl}. We also perform the experiments with two different object detectors. Results suggest that each cue contributes to improve the tracking performance. 

We experienced the aforementioned collapsed model issue when training the motion model on nuScenes without the bounding box size information. To overcome this, we first warmed up the network with the batch all online mining strategy and then we trained it with the batch hard strategy. We did not suffer from this issue when using the bounding box size as additional input, since the box size assists the network to discriminate too hard examples.

\begin{table}[]
\resizebox{\columnwidth}{!}{\begin{tabular}{lccccc}
\bottomrule
Detector                   & Mot          & App          & Boxs         & AMOTA $\uparrow$ & AMOTP $\downarrow$ \\
\bottomrule
\multirow{4}{*}{QD-3DT~\cite{Hu2021QD3DT}}    & -            & $\checkmark$ & $\checkmark$ & $0.235$            & $1.512$              \\
                           & $\checkmark$ & -            & -            & $0.285$          & $1.485$            \\
                           & $\checkmark$ & $\checkmark$ & -            & $0.289$          & $1.475$            \\
                           & $\checkmark$ & $\checkmark$ & $\checkmark$ & $\mathbf{0.295}$ & $\mathbf{1.468}$   \\
\bottomrule
\multirow{4}{*}{Mapillary~\cite{Simonelli_arXiv_2019}} & -            & $\checkmark$ & $\checkmark$ & $0.204$            & $1.515$              \\
                           & $\checkmark$ & -            & -            & $0.251$          & $1.489$            \\
                           & $\checkmark$ & $\checkmark$ & -            & $0.284$          & $1.474$            \\
                           & $\checkmark$ & $\checkmark$ & $\checkmark$ & $\mathbf{0.293}$ & $\mathbf{1.468}$  
\end{tabular}}
\caption{\textbf{Effect of each cue.} The pipeline is evaluated on nuScenes validation set using Mapillary~\cite{Simonelli_arXiv_2019} and QD-3DT~\cite{Hu2021QD3DT} object detectors.}
\label{tab:nusc-val-abl}
\end{table}

%-------------------------------------------------------------------------

\subsection{Benchmark results}
\label{sec:benchmarks-res}

\textbf{nuScenes}. We report the quantitative results of our method in Table~\ref{tab:nusc-test}, obtained on the test set.  The table shows that our method TripletTrack outperforms prior work by a large margin $23.5\%$. Compared to state-of-the-art methods our pipeline improves the tracking score relative to each object category and nearly all submetrics, with a huge reduction of identity switches. Contrarily to the current leading method QD-3DT~\cite{Hu2021QD3DT}, our pipeline makes use of keyframes only. This means that our method achieves higher performance while operating at a lower frame rate.

\textbf{KITTI \ac{mot}}. We report the quantitative results in Table~\ref{tab:kitti-test-car} and Table~\ref{tab:kitti-test-ped} obtained on the test set. Since our method benefits from more data, we pre-train our models on nuScenes and we fine-tune them on KITTI. Our method is competitive with most existing ones. Note that some of those performing better need to jointly train the object detector with the tracking mechanism~\cite{Chaabane2021deft} or make use of simulation platforms to generate synthetic training data~\cite{Tokmakov_2021_ICCV}. By looking at the HOTA metric components we can notice that the gap between our tracking pipeline and the best methods is higher on the detection submetrics compared to the ones related to tracking. This suggests that our pipeline compares well with top methods on data association and that if it were paired with a better object detector it would produce even higher overall tracking scores. Since the proposed pipeline gets the highest AssPr score, it sets a new Pareto optimal front on the submetrics pair AssPr-AssRe for camera-only tracking methods. This also confirms that our method is able to keep the number of identity switches low.

\begin{figure}[h!]
    \centering
    \includegraphics[width=\linewidth, trim={0 {5.5cm} 0 {10cm}}, clip]{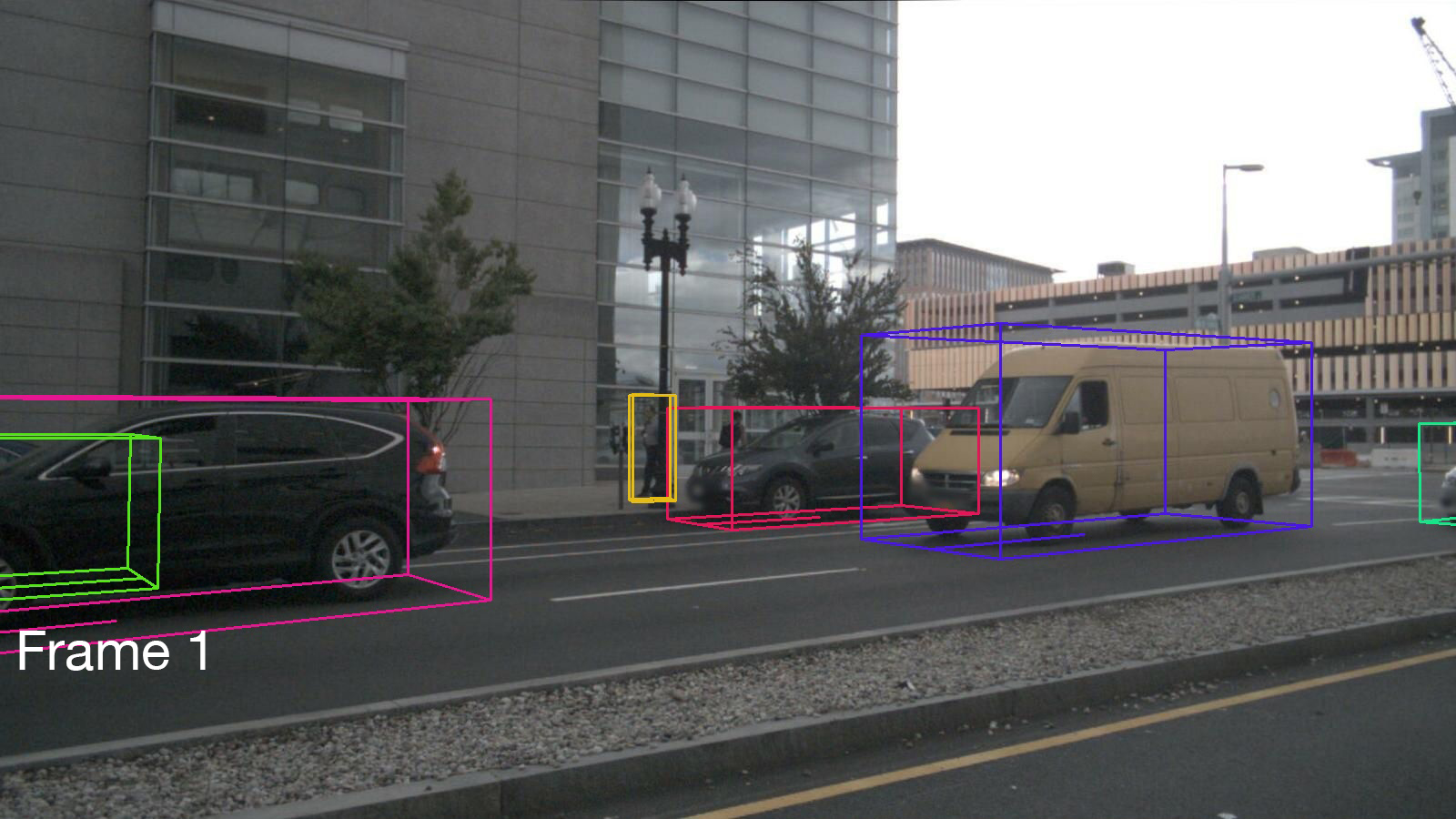}
    \includegraphics[width=\linewidth, trim={0 {5.5cm} 0 {10cm}}, clip]{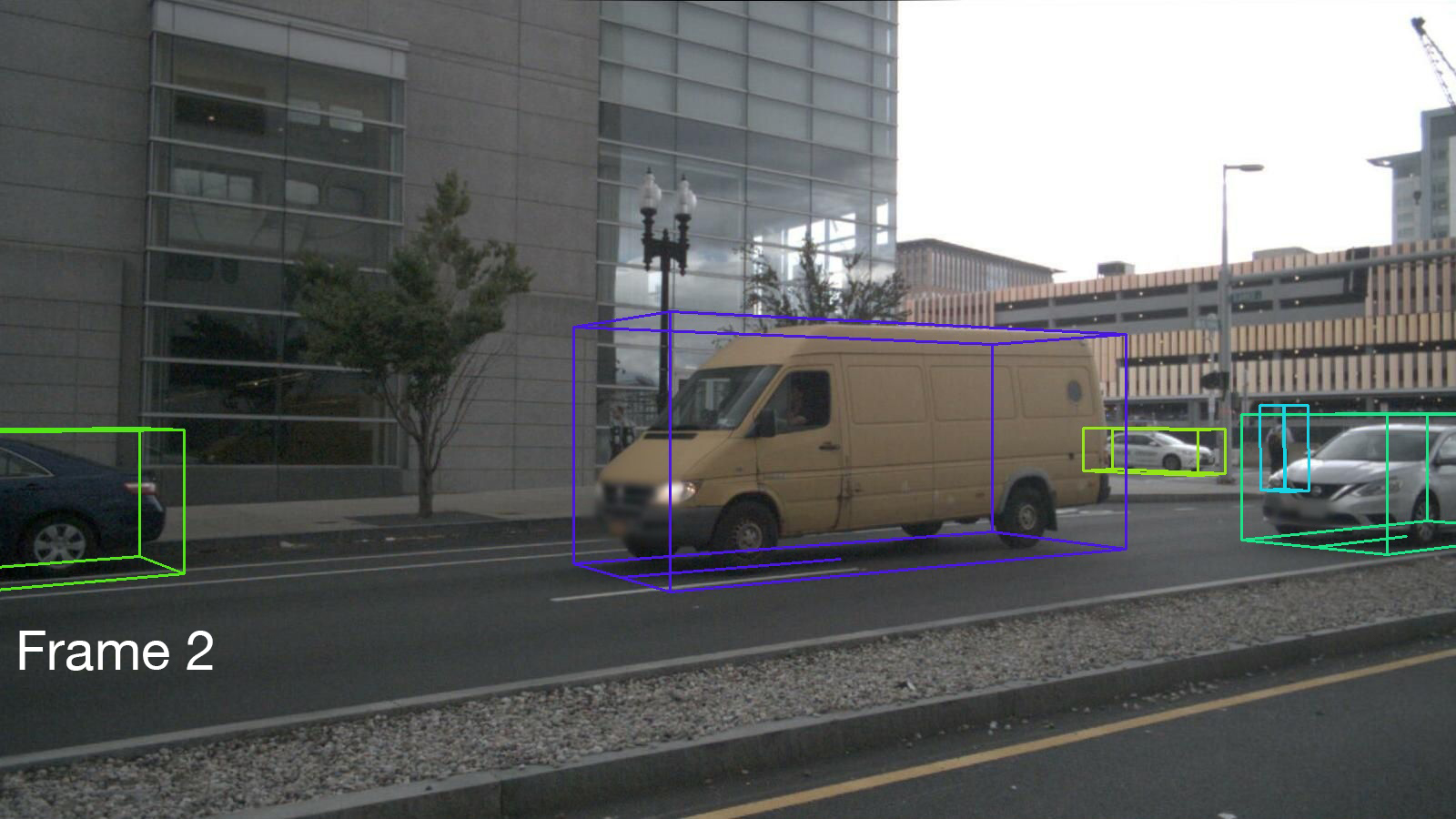}
    \smallskip
    \includegraphics[width=\linewidth, trim={0 {5.5cm} 0 {10cm}}, clip]{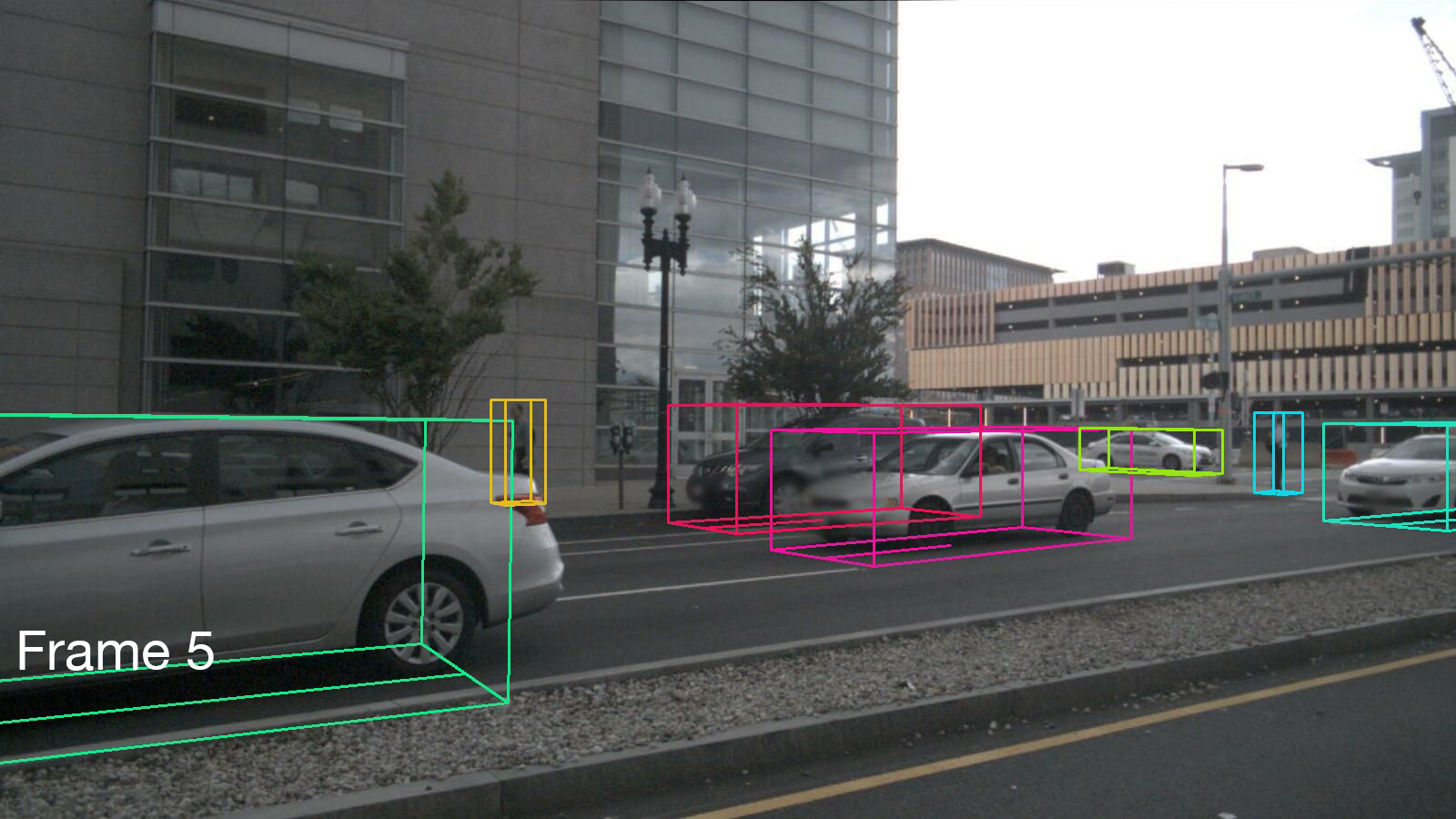}
    \includegraphics[width=\linewidth, trim={0 {5.5cm} 0 {10cm}}, clip]{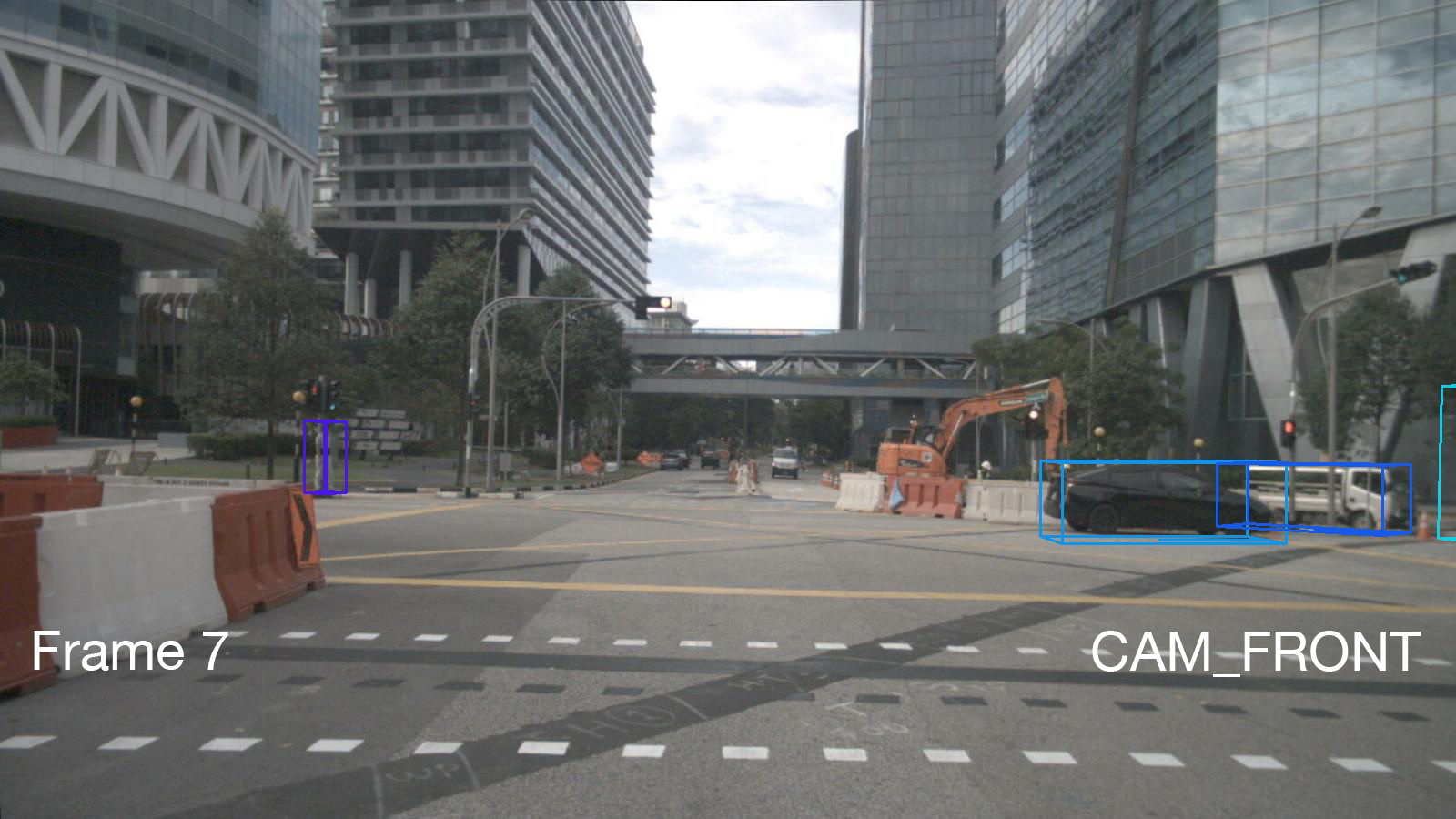}
    \includegraphics[width=\linewidth, trim={0 {5.5cm} 0 {10cm}}, clip]{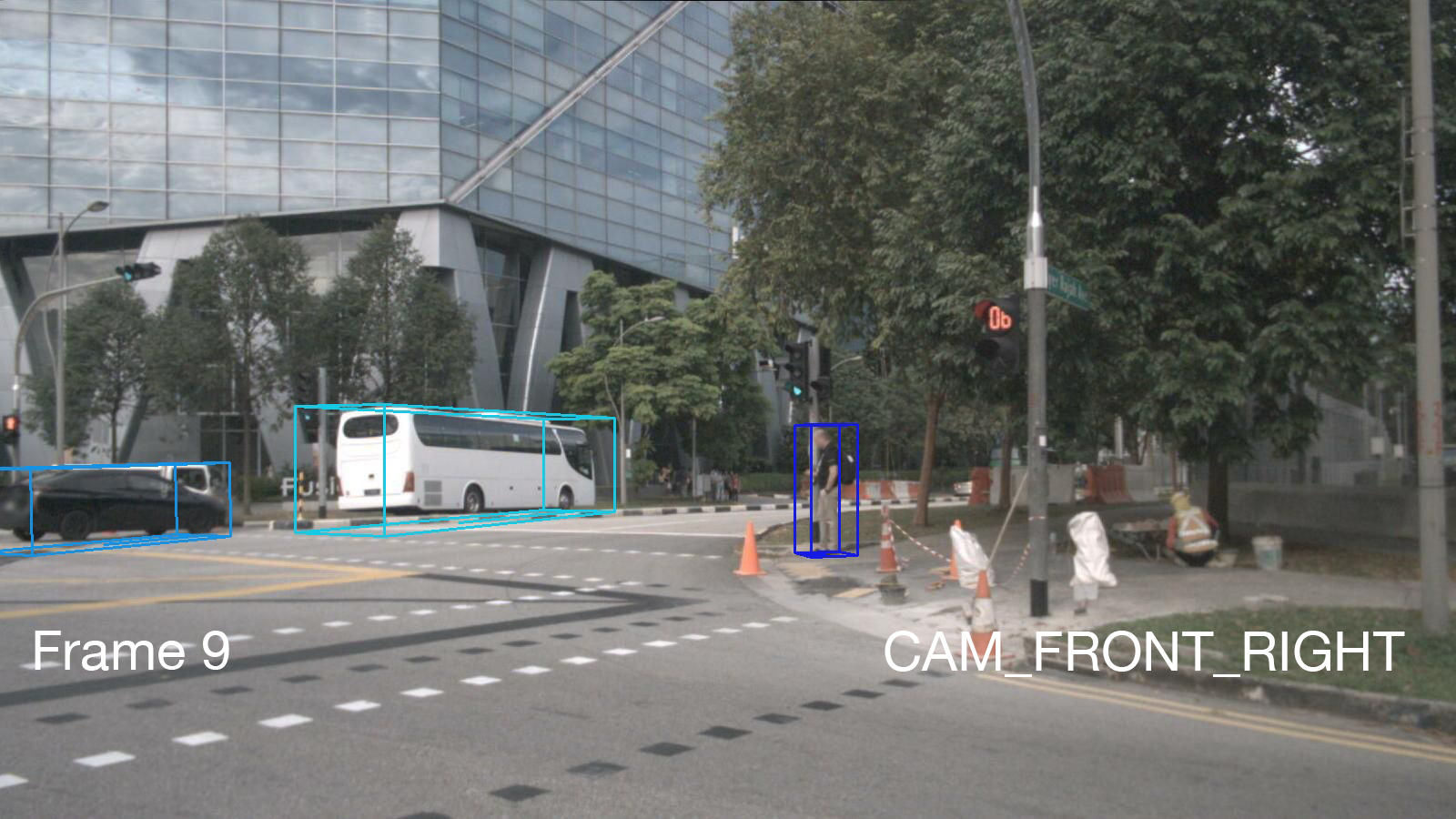}
    \includegraphics[width=\linewidth, trim={0 {5.5cm} 0 {10cm}}, clip]{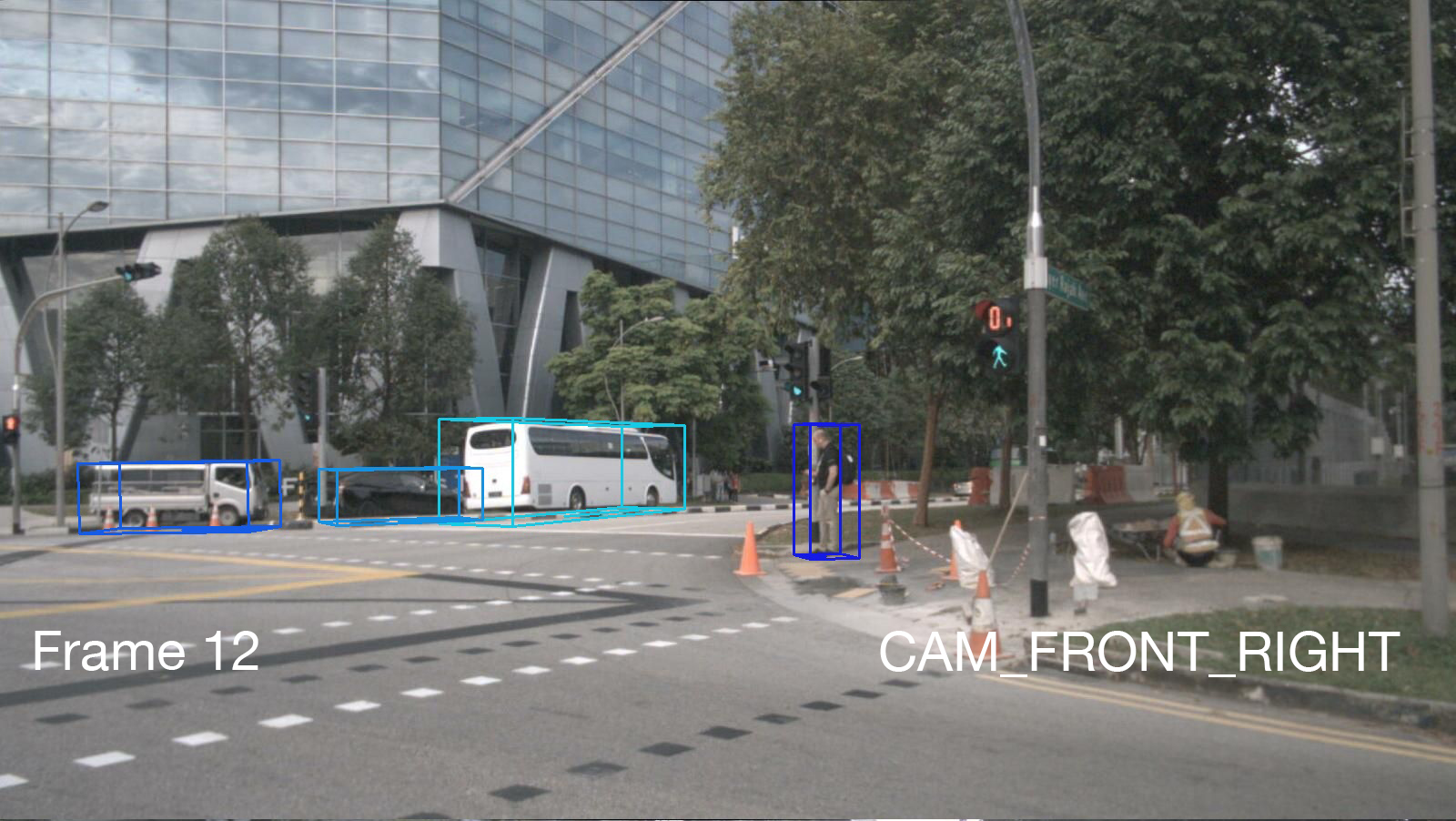}
    \caption{Top: illustration of pedestrian (yellow) and car (red) being occluded and picked up again. Bottom: Illustration of a car and truck re-identified across different cameras.
    }
    \label{fig:examples}
\end{figure}

In Figure~\ref{fig:examples}, we show two additional examples of our re-identification pipeline, dealing with occlusion between different objects and camera viewpoints.

\section{Conclusions}
\label{sec:conclusions}

We propose TripletTrack, a camera-only 3D multi object tracking method that combines appearance features and motion information. Appearance features are extracted through a \ac{cnn} trained with the triplet loss while motion data is processed by an \ac{lstm}. The ID assignment step is performed by the Hungarian algorithm, based on an affinity matrix. The affinity matrix is computed by an affinity network that combines appearance features and motion representation. Our method can also work on top of any 3D object detector and camera setup. Finally it outperforms the camera-based state-of-the-art on the nuScenes tracking benchmark and obtains competitive results on KITTI \ac{mot}.

\textbf{Acknowledgment.} The authors thankfully acknowledge support by Toyota via the TRACE project. Furthermore, we would like to thank Jonas Heylen and Bruno Dawagne for their valuable insights and comments.

%-------------------------------------------------------------------------

%%%%%%%%% REFERENCES
{\small
\bibliographystyle{ieee_fullname}
\bibliography{triplet_tracking}
}

\end{document}